\documentclass[letterpaper, conference]{IEEEtran}
\IEEEoverridecommandlockouts
\usepackage{cite}
\usepackage{amsmath,amssymb,amsfonts}
\usepackage{algorithmic}
\usepackage{graphicx}
\usepackage{textcomp}
\usepackage{hyperref}
\usepackage{xcolor}
\usepackage[letterpaper,margin=0.75in]{geometry}

\def\BibTeX{{\rm B\kern-.05em{\sc i\kern-.025em b}\kern-.08em
    T\kern-.1667em\lower.7ex\hbox{E}\kern-.125emX}}
\begin{document}

\title{\vspace{6mm}Tracking Emotions: Intrinsic Motivation Grounded on Multi-Level Prediction Error Dynamics
\thanks{GS has received funding from the European Union’s Horizon 2020 research and innovation programme under the Marie Sklodowska-Curie grant agreement No. 838861 (Predictive Robots). Predictive Robots is an associated project of the Deutsche Forschungsgemeinschaft (DFG, German Research Foundation) Priority Programme "The Active Self".
GS has also received funding from the EU-H2020 Open Cloud Research Environment (OCRE) project and the Marie Curie Alumni Association as part of the European Open Science Cloud initiative.
}
}

\author{\IEEEauthorblockN{Guido Schillaci}
\IEEEauthorblockA{\textit{The BioRobotics Institute} and \\
\textit{Dept. of Excellence in Robotics \& AI}\\
\textit{Scuola Superiore Sant’Anna}\\
Pisa, Italy \\
guido.schillaci@santannapisa.it}
\and
\IEEEauthorblockN{Alejandra Ciria}
\IEEEauthorblockA{\textit{Facultad de Psicolog\'ia} \\
\textit{Univ. Nacional Aut\'onoma de M\'exico}\\
M\'exico City, M\'exico\\
ciriacontact@gmail.com}
\and
\IEEEauthorblockN{Bruno Lara}
\IEEEauthorblockA{\textit{Centro de Investigaci\'on en Ciencias} \\
\textit{Univ. Aut\'onoma del Estado de Morelos}\\
Cuernavaca, M\'exico \\
bruno.lara@uaem.mx}
}

\maketitle

\begin{abstract}
How do cognitive agents decide which is the relevant information to learn and how goals are selected to gain this knowledge? Cognitive agents need to be motivated to perform any action. We discuss that emotions arise when differences between expected and actual rates of progress toward a goal are experienced. Therefore, the tracking of prediction error dynamics has a tight relationship with emotions. 
 Here, we suggest that the tracking of prediction error dynamics allows an artificial agent to be intrinsically motivated to seek new experiences but constrained to those that generate reducible prediction error. We present an intrinsic motivation architecture that generates behaviors towards self-generated and dynamic goals and that regulates goal selection and the balance between exploitation and exploration through multi-level monitoring of prediction error dynamics. This new architecture modulates exploration noise and leverages computational resources according to the dynamics of the overall performance of the learning system. Additionally, it establishes a possible solution to the temporal dynamics of goal selection. The results of the experiments presented here suggest that this architecture outperforms intrinsic motivation approaches where exploratory noise and goals are fixed and a greedy strategy is applied.
\end{abstract}


\section{Introduction}

A cognitive system can be conceived as one which fulfills its goals anticipating the causes of its sensations by containing a predictive model of itself and its environment to select and guide action. In Cognitive Robotics, a relevant question is how artificial agents should learn their internal models in order to interact efficiently with the world. Additionally, how can an artificial agent autonomously select a goal to pursue and decide what is the relevant information to learn during its interactions with the world? Here, the main research interest is the study of prediction error dynamics as the key element of an intrinsic motivation mechanism. Prediction error is the difference between the predicted sensory input an agent does and the sensory input from the world. Special interest lays on the impact these dynamics bring into the learning capabilities and interaction with the world of an artificial agent. Taking inspiration on the relevance of predictions on behavior and the relevance of the prediction error dynamics on learning, we suggest a new intrinsic motivation architecture. 

Internal models are acquired through the interaction of the agent with the world by learning the regularities present in the environment and on its internal states. This learning implies acquiring a sensorimotor schema which maps specific actions to its sensory consequences. Traditionally, internal models are composed by coupled \textit{inverse} and \textit{forward} models. The inverse model is a controller, which generates the motor command required to achieve a desired sensory state given a current sensory state. The forward model generates a predicted sensory state given a current state and the motor command provided by the inverse model. These models allow the prediction of the incoming sensory information and are constantly improved by monitoring unexpected states which are further processed. 

In developmental robotics, motor-babbling and goal-babbling are two commonly used approaches that allow artificial agents to autonomously and progressively learn a sensorimotor schema \cite{rolf2014autonomous, jamone2011learning,rayyes2019online, schillaci2016exploration}. Learning the sensorimotor consequences of self-generated actions during active interactions with the environment should progressively lead to a better model of itself. An agent that constantly improves the model of itself in the world should be better in avoiding surprising states. Minimizing prediction error leads to a better model of itself due to more accurate predictions about its future states. However, cognitive agents continuously search for novelty and unpredictable states. Paradoxically, predicting future states accurately while seeking unanticipated novel states are competing pressures within an agent. Here, it is suggested that these competing pressures should be resolved by tracking the dynamics of prediction error over time. In this line of thought, agents seek novelty but that which appears constrained to situations that generate reducible prediction errors. To do so, agents rely on their capacity to understand the unexpected information given the current internal models’ knowledge and the actions it can rely on \cite{van2017affective}. Control theories of self-regulation highlight the role of discrepancies and its velocities, and the affective response to this discrepancies in goal-directed behavior (e.g., \cite{carver1990origins, lord1987control}). We believe that the tracking of prediction error dynamics could be considered as a self-regulation mechanism that allows an agent to be intrinsically motivated to increase the complexity of behavior but constrained to seeking new knowledge that generates reducible prediction error. 

“Motivation can be operationalized in terms of goals, which are distributed and multifactorial, including biologically and affectively salient outcomes, sensory-motor plans, and integrated utility representations” \cite[p.1]{o2020unraveling}. Motivation is necessary to perform an action and to promote development \cite{von2003development}. Different computational models have been proposed inspired on intrinsic motivation to generate spontaneous exploratory behaviors and curiosity \cite{oudeyer2009intrinsic}. “Approaches using intrinsic motivation are crucial for the self-generation of goals, leading to an empirical process of exploration and the progressive acquisition of increasingly complex skills in a continual fashion” \cite[p.2]{parisi2019rethinking}. Existing computational approaches for task-independent learning and artificial curiosity are typically based on artificial agents that learn to anticipate the consequences of their actions and which actively can choose actions according to some internal measures related to the novelty or predictability of the anticipated situation (e.g. \cite{oudeyer2007intrinsic, schillaci2020intrinsic}). 

Here, we contribute to this line of research by proposing an intrinsic motivation architecture that generates exploratory behaviors towards self-generated goals, and that regulates goal selection and the balance between exploitation and exploration through a multi-level monitoring of prediction error dynamics. The proposed architecture implements a multi-level monitoring mechanism, which keeps track of the dynamics of the prediction errors at the goal level and of the dynamics of the overall error of the system at a general level. Moreover, the trend of the overall error at a general level drives the balance between exploitation and exploration. 
An agent sensitive to prediction error dynamics will be curious and motivated to explore its environment \cite{kiverstein2019feeling}. Similarly to other intrinsic motivation architectures, our approach 
drives behavior towards minimizing prediction error, i.e. towards goals associated with a steep descent in the prediction error dynamics.


There are two main differences between the intrinsic motivation architectures we are suggesting here and previous works in this field. First, it is based on the tracking of prediction error dynamics and its tight relationship with emotions. As \cite[p.5]{o2020unraveling} suggests: “The importance of these goal-oriented states for our everyday lives again points to the centrality of motivation for understanding human cognition, and a key computational modeling challenge is to construct the necessary metacognitive monitoring and motivationally significant grounding to enable a model to capture the corresponding goal dynamics.” Emotions represent the constant monitoring of ‘how things are going’ with respect to the expected progress by tracking the changes on prediction error during behavior \cite{carver1990origins}. An “emotional” artificial agent should be able to learn and autonomously select the proper prediction error reduction for any given situation. This capability allows an agent to valence an experience as positive or negative to itself, motivating the selection of the most suitable behavior for learning. Second, it establishes a possible solution to the temporal dynamics of goal selection. A relevant question for computational models is related to the temporal dynamics of when a goal is selected, for how long this goal is maintained, and how to know when a goal has been achieved in order to select another one \cite{o2020unraveling}. 

Emotions can be conceived as a set of valuating mechanisms that organize behavior \cite{adolphs2017should} prioritizing actions by the potential costs and benefits for an agent \cite{rushworth2004action}. This self-regulation mechanism, based on emotions, is intrinsically related to the needs of an agent and could be considered as the main performance regulator. Emotions are intrinsically related to goals, and they arise when differences between expected and actual rates of progress toward or away from those goals \cite{carver1990origins}. Therefore, it has been suggested that prediction error dynamics are the fundamental cause of emotions \cite{van2017affective}. Taking the assumption that biological agents have an innate system of positive and negative emotional valences, artificial emotions could be implemented by assigning emotional valences to specific characteristics of the prediction error dynamics. Thus, for an emotional experience, monitoring prediction error dynamics becomes equally relevant as the monitoring of predictive errors per se. 

There are various hypotheses of how emotional valence could be determined by the amount of change in prediction errors over time (e.g., \cite{carver1990origins, joffily2013emotional,van2017affective,van2011putting,kiverstein2019feeling}). For example, it has been suggested that a positive valence is linked to an active reduction of prediction errors, and a negative valence to a continuous increase of prediction errors \cite{van2017affective, van2011putting}. Another hypothesis is that the emotional valence depends on the expected rate of reduction of prediction error \cite{carver1990origins}. In this view, changes in the reduction rate of prediction error leads to a change on the emotional experience. This hypothesis implies that the reduction rate has to be learned in order to have a reference value that should be in an acceptable rate of behavioral discrepancy reduction. The rate of change, in relation to prediction error, is analogous to velocity because it refers to how fast or slow is being reduced relative to a frame of reference over time \cite{joffily2013emotional,carver1990origins}. “Each agent’s performance in reducing error can be plotted as a slope that depicts the speed at which errors are being accommodated relative to their expectations. The steepness of the slope indicates that error is being reduced over a shorter period of time and so faster than the agent expected: the steeper the slope, the faster the rate of reduction” \cite[p.2857]{kiverstein2019feeling}.

How does an emotional experience based on the tracking of prediction error dynamics enhance learning? Prediction error and its reduction rates might signal the expectations on the learnability of particular situations \cite{van2017affective}, and could potentially guide attention to prioritize learning of the most informative input domain. In the rise of affect, “the discrepancy that matters is a discrepancy in sensed progress towards ideals” \cite[p.32]{carver1990origins}. This information means that the progress of an action toward its goal is high. If the speed of reduction increases relative to what was expected, or even faster, a positive emotional experience should emerge. Contrary, when the prediction error does not decrease in an expected way, the task becomes ``uncomfortable'' or ``frustrating''. This ``continuous unpleasant surprise'' should induce the disengagement from the activity in order to seek another goal that enhances learning. These ideas are related to the dynamic approach to outcome-satisfaction relations suggested by \cite{hsee1991velocity}. They proposed that what motivates an engagement in a behavior is not just the final outcome, but the satisfaction that emerges from the pattern and the velocity of an outcome over time.

Regarding the open questions and issues in the computational modeling frameworks about the temporal dynamics of goals \cite{o2020unraveling}, we believe that tracking the prediction error dynamics over time could provide useful answers. First, concerning the question about when a goal is selected, we have suggested that the drive to minimize prediction error should motivate behavior and thus a goal-selection just when a previous goal has been achieved or abandoned. Minimizing prediction errors evokes a positive emotional experience, motivating the system to constantly move into a goal-selection phase in order to initiate a goal-engaged state. Second, tracking the prediction error dynamics during the execution of a goal allows a direct monitoring of the progress towards its achievement. Here, we propose that a goal is accomplished when the steepness of the slope indicates that error is being continuously reduced until reaching the desired outcome. 

A relevant question here is which is the optimal size of the time window that prediction errors should be monitored. Progress towards, and maintenance of a goal result in an affective reaction and influence the way available resources are used \cite{kluger1996effects}. Taking inspiration of the control theory of human motivation proposed by \cite{lord1987control}, large prediction errors should elicit a more careful evaluation of feedback to verify if prediction errors are actually increasing. This careful evaluation should permit a fast correction of behavior or induce a disengagement of the goal to avoid frustration. On the other hand, we believe that when a goal is well performed the necessity of tracking the prediction errors can be reduced, liberating processing resources, which can give a potential explanation of how habits and automatic behaviors emerge. Therefore, we suggest that the size of the time window of where prediction errors are tracked should not be monotonic or fixed. The intrinsic motivation architecture proposed here dynamically adjust the size of the time window in response of ‘how things are going’ with respect to the expected progress.





\section{Methodology}

\begin{figure*}[htbp]
\centerline{\includegraphics[width=12cm]{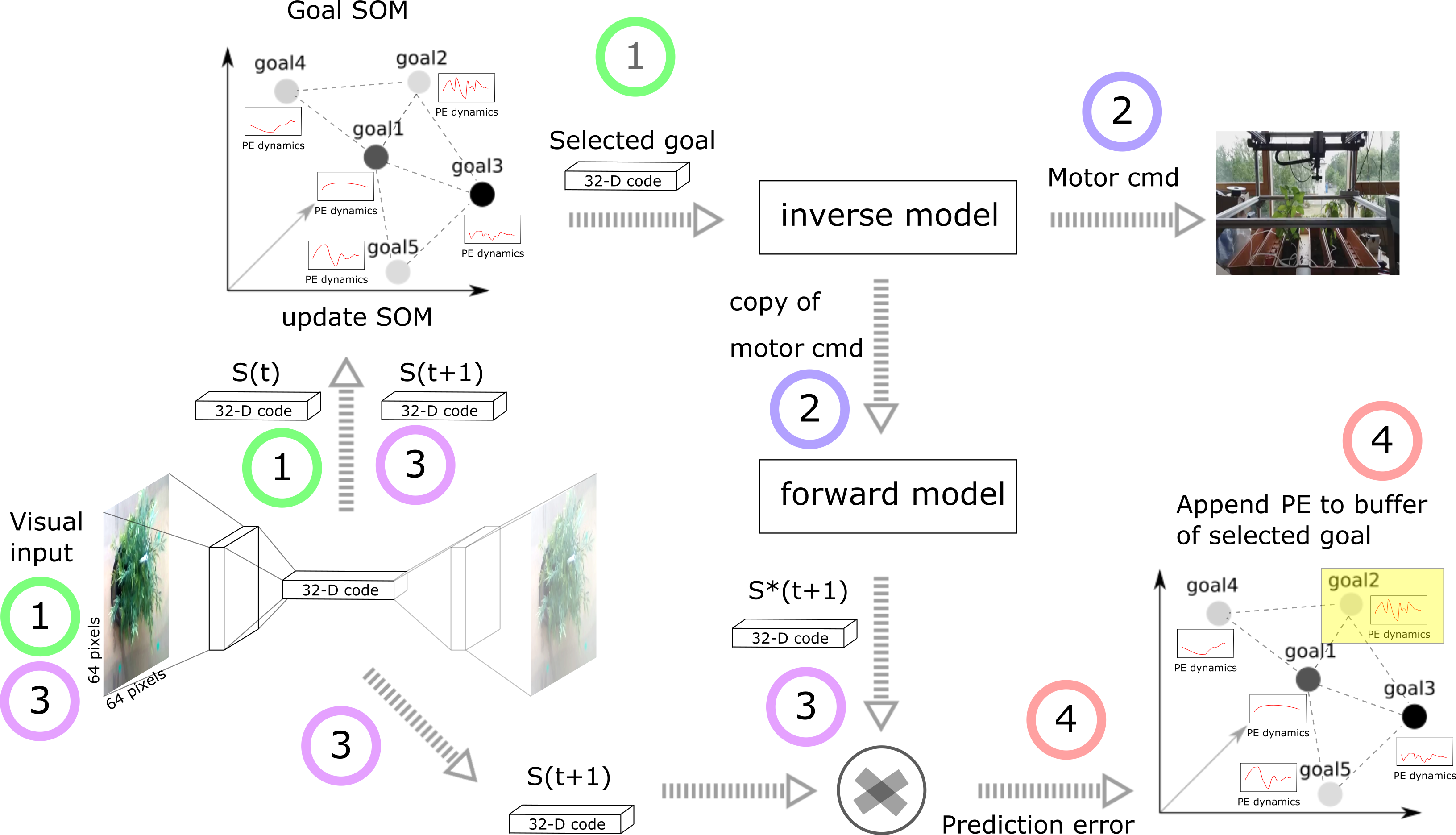}}
\caption{The learning architecture. Numbers indicates the order at which information is processed.}
\label{fig:architecture}
\end{figure*}

We propose a learning architecture that generates exploratory behaviours towards self-generated goals, and that regulates goal selection and the balance between exploitation and exploration through a multi-level monitoring of prediction error dynamics.
The proposed architecture implements a multi-level monitoring mechanism, which keeps track of the dynamics of the prediction errors at the \textit{goal level} and of the dynamics of the overall error of the system at a \textit{general level}. The multi-level monitoring process is described in section \ref{sec:method:pe_monitoring}.
The trends of the prediction error dynamics at the \textit{goal level} drive the goal selection strategy, as described in section \ref{sec:method:goal_selection}. Moreover, the  trend of the overall error at a \textit{general level} drives the balance between exploitation and exploration. This is implemented by modulating the amount of noise added to goal-directed movements (see section \ref{sec:method:expl_expl}). 

The proposed architecture is tested on a simulated experiment where a robot has to learn visuo-motor coordination. The system is characterised by high-dimensional visual inputs captured by a camera whose movements are controlled by a two degrees-of-freedom actuator (see section \ref{sec:results}).

The architecture, depicted in Figure \ref{fig:architecture}, is composed by four neural networks: (1) a deep convolutional autoencoder (CAE) which is in charge of reducing the dimensionality of the visual input (see section \ref{sec:method:cae}); (2) a self-organising map \cite{kohonen1998self} for unsupervised learning of visual goals, hereon named \textit{goal SOM} (see section \ref{sec:method:goals}); (3) a deep neural network representing the controller, or \textit{inverse model}, of the system, which learns the motor commands required to reach the generated goals; (4) a deep neural network representing the predictor, or \textit{forward model}, of the system, which learns to anticipate the visual input that would be captured after the execution of a specific motor command. Section \ref{sec:method:inv_fwd} describes the inverse and forward models in more detail.

\subsection{Unsupervised learning of features from visual inputs}
\label{sec:method:cae}

In the experiments presented here, visual inputs consist of $64 \times 64$ pixels grayscale images captured from a robot camera.
A convolutional autoencoder (CAE) is used for unsupervised learning of low-dimensional features from such high-dimensional data, typically images \cite{masci2011stacked}. Typically, CAEs are characterised by a hidden layer, the \textit{latent} layer, of considerably lower dimensionality than their input. The information propagated from the input layer to the latent layer of a trained CAE can be interpreted as an encoded, or compressed, version of the input \cite{masci2011stacked}. Here, a CAE is trained to reduce the dimensionality of the $64 \times 64$ visual inputs onto 32-dimensional codes, hereon named \textit{sensory states}, or simply \textit{S}. That is, visual inputs -- as well as visual goals -- will be encoded as 32D vectors. 

The dimensionality of the visual inputs is reduced to ease prediction error calculations.
Calculating the dissimilarity between a predicted image and an observed one is not trivial. In computer vision, dissimilarity between images is typically not estimated using pixelwise difference, but rather by comparing a set of features extracted from them. \cite{chen2005similarity}.

In this work, the CAE is pre-trained during a learning session preceding the series of experiments, as presented in section \ref{sec:results}.
Once trained, the weights of the CAE model are not updated anymore during the experiments.


\subsection{Unsupervised learning of visual goals}
\label{sec:method:goals}

The behaviours generated by our system are goal-directed\footnote{However, as it will be described in section \ref{sec:method:expl_expl} a mechanism that monitors the error dynamics at a general level is modulating the amount of exploratory noise that is added to the motor commands. Eventually, this can make a goal-directed movement becoming, in fact, a random movement.}. A self-organizing map (SOM) is used to autonomously generate goals for the goal-directed exploration behaviour. The \textit{goal SOM} is updated with the sensory states -- i.e. the 32D codes resulting from visual inputs compressed by the CAE -- that are observed by the artificial system throughout its lifetime.

The \textit{goal SOM} is characterised by a set of neurons that self-organise around the observed sensory states. Each region of the sensory space is represented by a \textit{goal SOM} neuron.
We consider goals as the 32D position of the neurons in the SOM's feature space.
Visual goals are therefore not pre-coded, rather learned using a  self-organising map training algorithm\footnote{The Minisom library (https://github.com/JustGlowing/minisom) has been re-adapted in the experiment presented here.}.


\subsection{Online learning of inverse and forward models }
\label{sec:method:inv_fwd}

Inverse and forward models are acquired through the interaction of the agent with the world. The inverse model is in charge of generating the necessary motor commands to achieve a goal passed as input. On the contrary, the forward model is in charge of predicting the sensory outcome of a given motor command that is passed as input. It has to be noted that, in the experiments presented here, the motor commands are absolute target motor positions, therefore no initial sensory state is needed as input in both internal models\footnote{This diverges from the classical view of internal models. We believe however that this has a limited conceptual impact on the architecture.}. 

The inverse and forward models are implemented as deep neural networks and updated in an online fashion with the data that is generated throughout the exploration behaviour of the artificial agent. In particular, the inverse model is a neural network that takes as input a 32D code (the goal). The input layer is followed by a series of dense and dropout layers\footnote{Dropout layers are used for reducing overfitting and improving the generalization of the network. A drop out rate of 10\% is used.}. The output layer is characterised by two neurons, representing the motor command of the two DoF robot.

The forward model takes as input a two-dimensional tensor, i.e. (the motor command. The input layer is followed by a series of dense and ReLU activation layers. Finally, a sense layer of 32 neurons with sigmoidal activation represents the output of the model -- i.e. the 32D sensory state. 
Both inverse and forward models are optimised on a mean squared error loss function using an AdaDelta optimiser.

Moreover, an episodic memory system has been adopted to prevent catastrophic forgetting issues, typically experienced when updating neural networks in an online fashion. A system utilised in a previous work \cite{schillaci2020intrinsic} has been adopted.

\subsection{Multi-level monitoring of prediction error dynamics}
\label{sec:method:pe_monitoring}

The proposed architecture regulates goal selection and the balance between exploration and exploitation based on a multi-level monitoring of prediction error dynamics. In particular, the framework monitors two types of error: (1) a high-level, \textit{general error} of the system, which consists of the mean squared error (MSE) of the \textit{forward model} calculated on a pre-recorded test dataset; (2) low-level goal errors, which consist of the prediction errors that are estimated by the system when trying to reach each specific goal. 


At the lower level, the framework maintains, for each goal, a buffer of prediction errors that have been estimated in the past. Every time a goal is selected, a movement directed to this goal is performed and a prediction error, calculated as the distance between the goal and the predicted sensory state $S^*(t+1)$ estimated using the forward model, is appended to this buffer. Each goal buffer has a variable size, which is dependent on the dynamics of the upper-level general error.

At the upper level, the system monitors the dynamics of the mean squared error of the \textit{forward model} calculated on a test dataset\footnote{Test dataset contains 200 visuo-motor samples randomly picked up from the simulator visuo-motor dataset, described in section \ref{sec:results}. Each sample represents a camera position, i.e. (x,y) motor position, and the corresponding image captured from the camera. The same test dataset is used for all the experiments presented in Section \ref{sec:results}.}. The system maintains a buffer of MSE observed during a specific time window. This buffer is updated at a lower pace than that of the lower-level goal buffers. An update of the MSE buffer consists of the following steps: for each sample in the test dataset, the system takes the values of the motor command and feeds them as input into the forward model; the predicted output is compared to the compressed visual input stored in the test sample. The resulting error is squared. The mean of the squared errors over all the test samples is calculated and appended to the MSE buffer.

After every update of the MSE buffer, a linear regression is calculated on the stored values over time. The slope of the linear regression indicates the trend of the general error of the system. This trend is used to modulate: (1) the size of the \textit{goal error buffers}; (2) the standard deviation of a Gaussian noise that is added to each motor command executed during the exploration (see section \ref{sec:method:expl_expl} for more details).

When the trend of the MSE error is positive -- i.e. the linear regression has a positive slope, indicating that performance of the system is worsening -- the size of the lower-level \textit{goal error buffers} is increased by one, until a maximum size is reached. On the contrary, when the trend of the general error is negative -- i.e. the linear regression has a negative slope, indicating an improvement of the overall performance -- the size of the lower-level \textit{goal error buffers} is decreased by one, until a minimum size is reached. In other words, when the system is experiencing a general improvement of the overall performances, the necessity of tracking the goal error dynamics is reduced. Therefore, the goal error dynamics are monitored over shorter time windows, driving exploration towards goals that bring \textit{instantaneous} reduction of the prediction error.
On the contrary, when a deterioration of the overall performances is experienced, the system increases the tracking of the goal error dynamics by widening the time window on which errors are monitored.

\begin{figure}[htbp]
\centerline{\includegraphics[width=7cm]{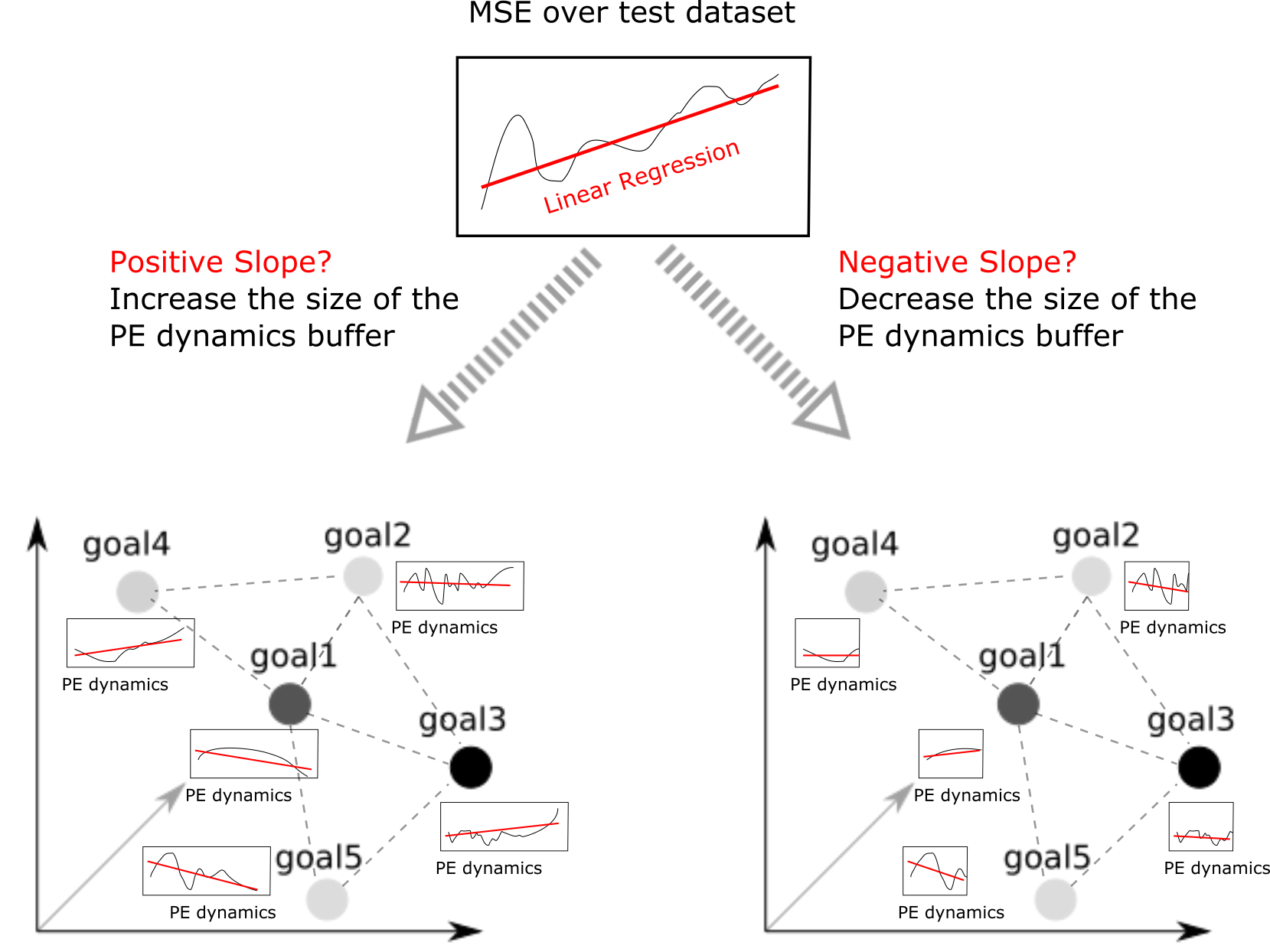}}
\caption{Multi-level monitoring of prediction error dynamics.}
\label{fig:dynamics}
\end{figure}

Figure \ref{fig:dynamics} shows an illustration of the multi-level monitoring process. 
Both the MSE and goal error buffers are fed with first-in-first-out (FIFO) strategy.

\subsection{Exploration vs. exploitation}
\label{sec:method:expl_expl}
The trend of the overall performance of the system controls also the amount of noise that is added to the exploration behaviours.
In fact, a random value sampled from a normal distribution with zero mean is added to the commands that are sent to the motors of the robot. The standard deviation of the normal distribution is set to be proportional to the slope of the linear regression of the MSE over time. The steepest the descent of the MSE, the smaller the standard deviation of the normal distribution and thus the noise. The smaller the noise, the closer the system explores the target goal (higher \textit{exploitation}). On the contrary, the steepest the ascent of the MSE, the higher the standard deviation of the normal distribution and thus the noise. When overall performances are deteriorating, the systems autonomously tends towards performing random exploration behaviours. Section \ref{sec:results} will illustrate some example of these behaviours and their impact on the overall learning progress.

\subsection{Goal selection strategy}
\label{sec:method:goal_selection}
Each goal is associated with an \textit{error buffer}. At every iteration of the exploratory behaviour, the system selects a goal and generates a goal-directed movement. Initially, a random goal is selected, i.e. a random neuron of the \textit{goal SOM} is selected and the corresponding position in the 32D sensory space is fed into the inverse model. The controller generates the best known motor command to achieve the selected goal.

After the execution of the movement, a prediction error is estimated as described in the previous sections, and appended to the error buffer of the \textit{currently selected goal}. When the buffer contains more than four estimated prediction errors, a linear regression is calculated on the stored values. The slope of the linear regression indicates the trend of the learning progress for the specific goal. A negative slope suggests that the activity directed at the current goal is bringing useful information that improves the learning progress. On the contrary, a positive slope suggests that the current activity is worsening the learning progress on the current goal.

The goal selection strategy chooses, at every iteration, the goal that is associated with the steepest descending trend of the prediction error dynamics. When the slope of the linear regression of the current goal becomes positive, or its absolute value smaller than a threshold, indicating that the current activity does not bring much improvement in terms of learning progress, then the system switches to a different goal. Again, the system gives priority to those goal that are associated with the steepest descent in the prediction error dynamics. To prevent that the system jumps between goals very quickly, when a goal is selected and the prediction error trend is not decreasing, the system maintains the goal for a minimum number of iterations (in the experiments, set to 50).
A \textit{greedy goal selection strategy}\footnote{Note: this is \textit{not} the greedy strategy generating random motor commands, as described in section \ref{sec:results}.} is also applied, i.e. the system chooses a random goal with a probability of 1.0\%.

\section{Results}
\label{sec:results}
The architecture is tested on a simulated  experimental setup, in which an artificial system has to learn visuo-motor coordination -- i.e. its inverse and forward models -- through a goal-directed exploration behaviour.
A simulated platform developed in a previous work has been adopted \cite{schillaci2020intrinsic}. The simulator consists of a script that generates trajectories of a camera of a micro-farming robot using a pre-recorded visuo-motor dataset \cite{guido_schillaci_2020_3552827}.
The robot consists of two motors, moving an RGB camera along a horizontal plane. The camera is facing top-down on a desk where some objects are located. The visuo-motor dataset has been generated by performing a full scan of the horizontal plane within the limits of the robot movements using a step of 5mm. This resulted in 24.964 images, each mapped to an (x,y) position of the motors. The simulator allows generating trajectories from a starting motor position to a target one, and returning the motor and visual samples observed along these trajectories.
The dataset is freely available \cite{guido_schillaci_2020_3552827}, as well as the source code behind this work\footnote{Repository: \url{https://github.com/guidoschillaci/prediction_error_dynamics}}.

The convolutional autoencoder described in section \ref{sec:method:cae} has been trained on the $24 964$ images stored in the visuo-motor dataset over 40 epochs. 
Once having trained the CAE, we carried out a set of experiments in which the system generated goal-directed behaviours using self-generated and self-regulated goals as described above. During the exploration behaviours, the system learned its inverse and forward models in an online fashion. We carried out experiments varying the following parameters: 

\begin{itemize}
    \item[1.] Whether the \textit{goal SOM} is pre-trained or is being trained during the experiments, resulting in either fixed or moving goals, respectively.
    \item[2.] Whether the exploration noise is adaptive to the trend of the MSE dynamics (see section \ref{sec:method:expl_expl}) or fixed.
    \item[3.] Whether a greedy exploration strategy is applied (with a probability of 3\%) or not. 
\end{itemize} 

Table \ref{tab:doe} illustrates the series of experiments resulting from the combination of these configurations. Every experiment has been run five times, each consisting of 5000 iterations. 

\begin{table}[h]
\begin{center}
\begin{tabular}{ |p{0.9cm}|p{1.9cm}|p{2cm}|p{2.2cm}|  }
 \hline
 \multicolumn{4}{|c|}{Design of experiments} \\
 \hline
  \hline
 Exp. ID & Fixed goal SOM  & Fixed expl. noise & Greedy expl. prob. \\
 \hline
  \hline
 0  & False & False & 0\% \\
 1  & True & False & 0\% \\
 2  & False & True & 0\% \\
 3  & True & True & 0\% \\
 4  & False & False & 3\% \\
 5  & True & False & 3\% \\
 6  & False & True & 3\% \\
 7  & True & True & 3\% \\
 \hline
\end{tabular}
\vspace{0.2cm}
\caption{}
\label{tab:doe}
\end{center}
\end{table}

In each iteration, movements generate visuo-motor samples consisting of paired (x,y) motor positions and $64 \times 64$ pixels grayscale images. Images are compressed into 32D sensory states using the CAE. A SOM with $3\times3$ neurons is adopted, thus encoding \textit{nine} goals. The SOM has a 32D feature space, as the sensory space.
Inverse and forward models are updated in an online fashion, whenever a batch of 16 visuo-motor samples is observed. When a batch is available, it is used together with the current content of the episodic memory as training data for the models update\footnote{
Keras deep learning library on TensorFlow backend have been used to build and train all the models presented here (except for the SOM). Simulations have been run using cloud services for GPU accelerated computing. In particular, Amazon Web Services (g4dn and p3) and Exoscale (GPU2) instances have been used, gently supported by the Marie Curie Alumni Association (MCAA) and the EU-H2020 Open Clouds for Research Environment projects.}. 
Maximum memory size is set to 1000 visuo-motor samples, with an update probability of 1\%. When full memory size is reached, only one \textit{random} sample in the memory is replaced with the current observation.

MSE dynamics are updated at a lower frequency than to the lower-level goal error dynamics. In particular,
mean squared errors of both the forward and inverse models are computed and logged every 40 iterations. The MSE of the inverse model is logged for performance analysis, whereas the MSE of the forward model is buffered and monitored, as discussed in the previous sections. The size of the MSE buffer is set to 10. A linear regression over the values stored in the buffer is computed at every update, so as to estimate the trend of the overall system performances. The size of the lower-level goal error buffers is initialised to 10 and, as described in section \ref{sec:method:pe_monitoring}, is modulated according to the trend of the MSE dynamics. The minimum and maximum sizes of the goal error buffer are set to 10 and 50 time steps, respectively.

\begin{figure}[htbp]
\centerline{\includegraphics[width=7.5cm]{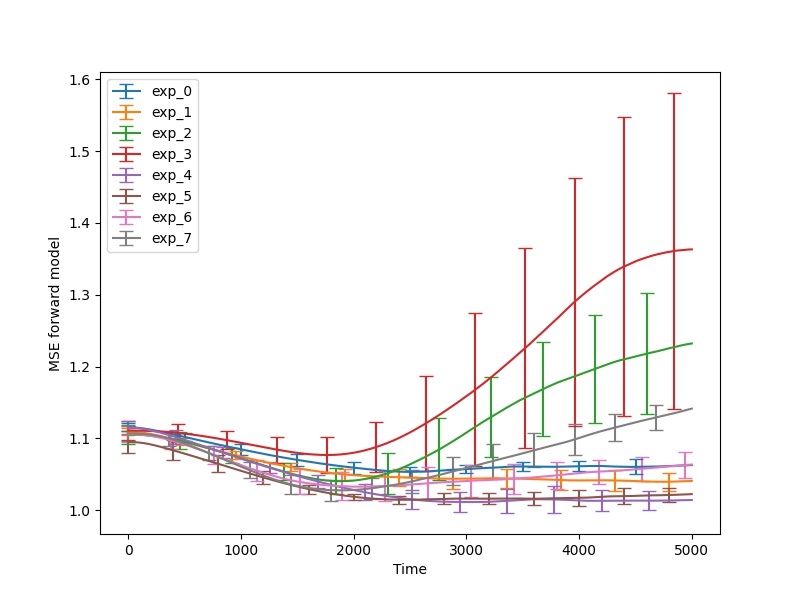}}
\caption{Mean squared error of the forward model for each experiment. Curves show MSE averaged over 5 runs per experiment.}
\label{fig:mse_fwd}
\end{figure}

\begin{figure}[htbp]
\centerline{\includegraphics[width=7.5cm]{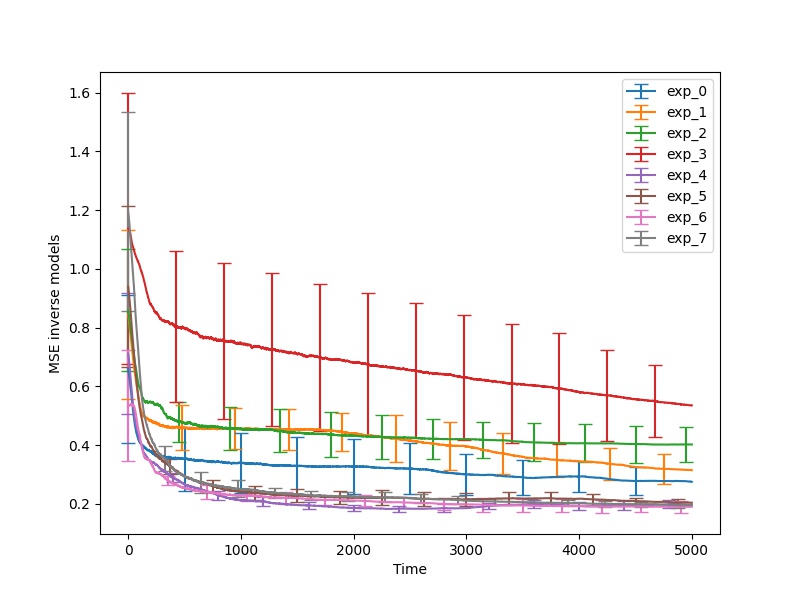}}
\caption{Mean squared error of the inverse model for each experiment. Curves show MSE averaged over 5 runs per experiment.}
\label{fig:mse_inv}
\end{figure}

Figures \ref{fig:mse_fwd} and \ref{fig:mse_inv} show the MSE of the forward and inverse models for each experiment, as in Table \ref{tab:doe}. Each curve represents the average MSE over 5 runs of the same experiment. We expect experiment \#3 to be the worst in terms of adaptivity and performances, considered that  exploratory noise and \textit{goal SOM} are both fixed, and no greedy strategy is applied. Results confirm our expectations. In both Figures \ref{fig:mse_fwd} and \ref{fig:mse_inv}, experiment \#3 has the slower descending learning progress. As for the forward model, performance is even deteriorating from the initial state as it can be seen from the ascending MSE red curve in Figure \ref{fig:mse_fwd}.
Similar performance, although slightly better, can be observed for experiment \#2 (light green curve), which differs from \#3 only in the online training of the \textit{goal SOM}. Fixing the exploratory noise to an initial small value and not adopting any greedy exploration strategy is likely not preventing the system to stuck in local minima.

The most interesting experiments are \#0 and \#4, in which goal locations are being learned over time and exploration noise is adaptive and dependent on the MSE dynamics. Experiment \#4, as it can be noted in Figures \ref{fig:mse_fwd} and \ref{fig:mse_inv}, outperforms all the other experiments. This confirms the well-known positive contribution of greedy strategies in intrinsic motivation systems. Experiment \#0 (light blue curve) presents also descending MSE trends for both inverse and forward models. Although not performing as well as experiment \#4, experiment \#0 indicates that the self-regulatory mechanisms that modulate exploration noise and dynamic goals can compensate for the absence of a greedy exploration strategy.
Both the experiments outperform standard intrinsic motivation approaches, where exploratory noise and goals are fixed, and greedy strategy is applied -- i.e. experiment \#7.

\begin{figure}[htbp]
\centerline{\includegraphics[width=4.5cm]{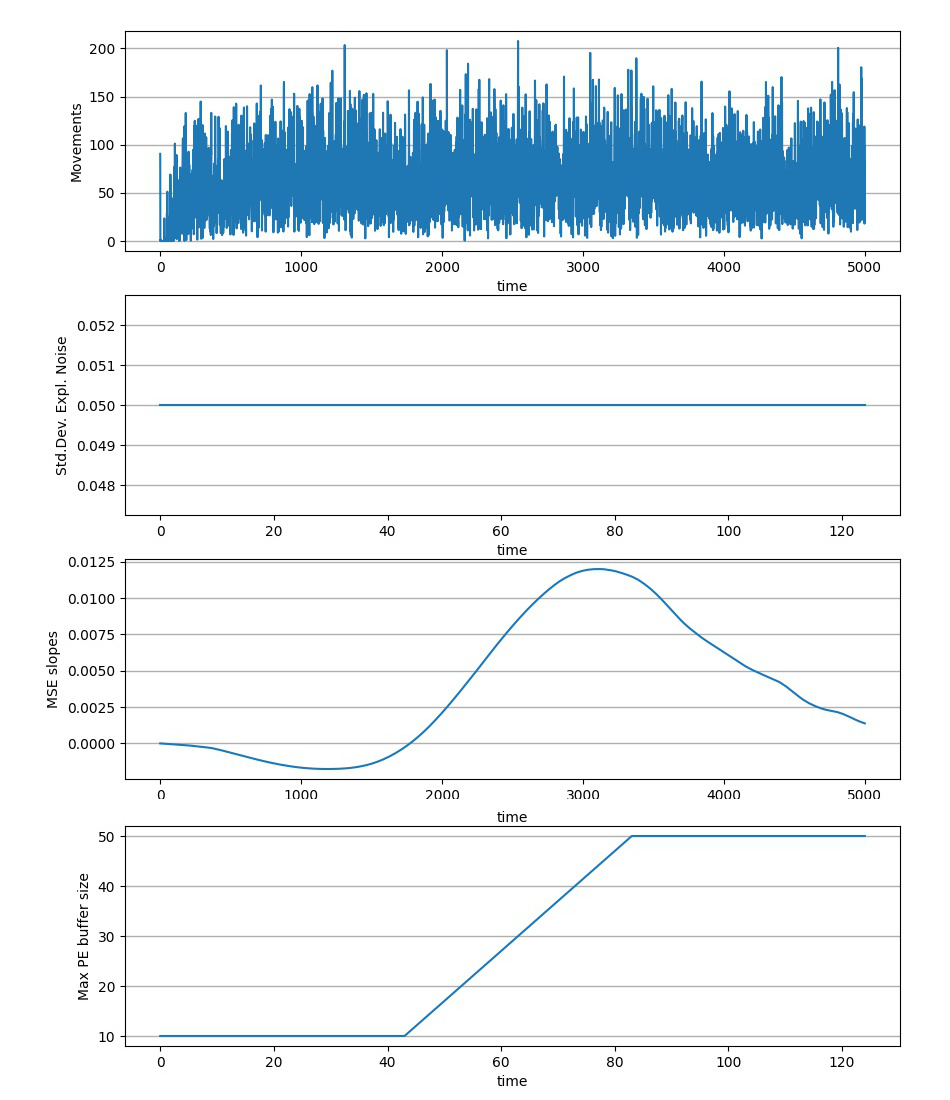}
\includegraphics[width=4.5cm]{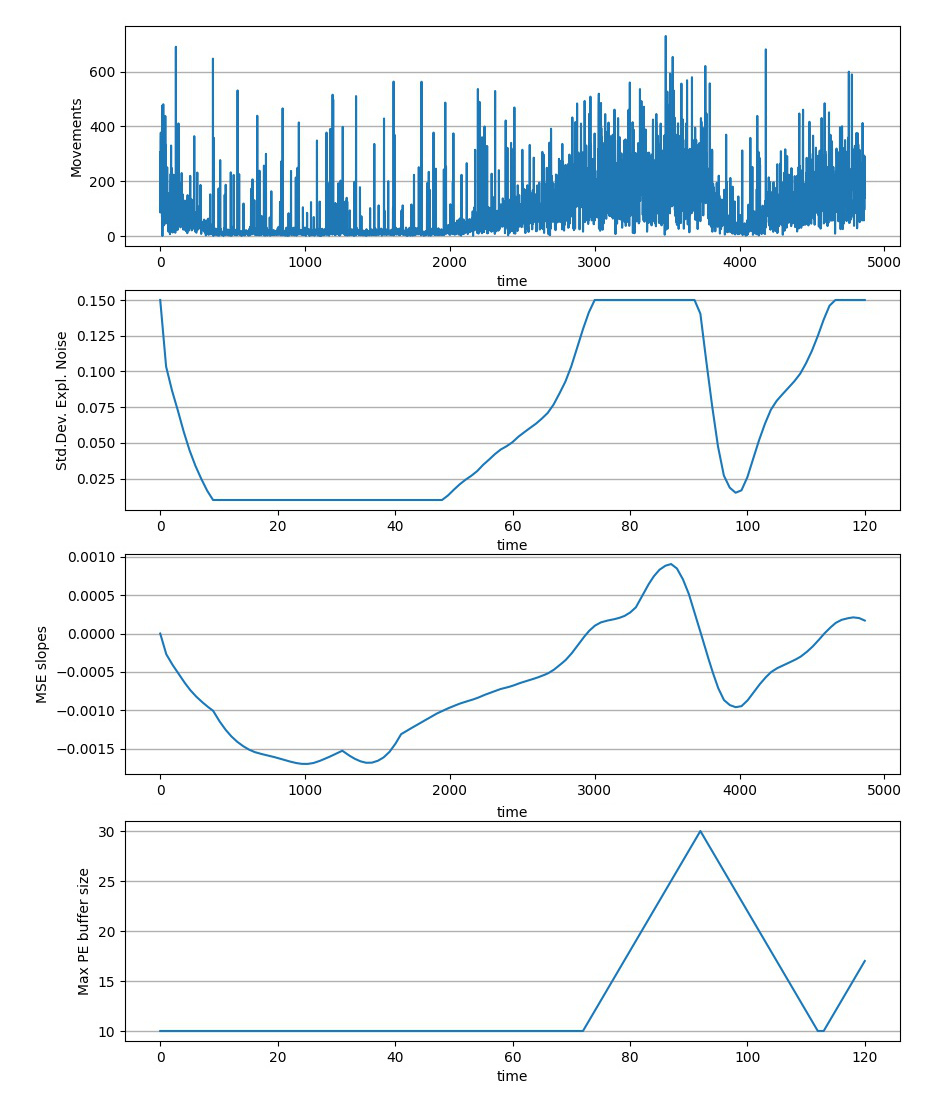}}
\caption{Detailed dynamics of two specific runs for experiment \#3 (left) and experiment \#4. Top plots show the amplitude of the robot movements. The second plots show the standard deviation of the normal distribution where the exploration noise is sampled from. The third plots show the slope of the linear regression calculated on the MSE buffer. The bottom plots show the dynamics of the maximum size of the goal error buffers.}
\label{fig:slopes}
\end{figure}

\begin{figure}[htbp]
\centerline{\includegraphics[width=9cm]{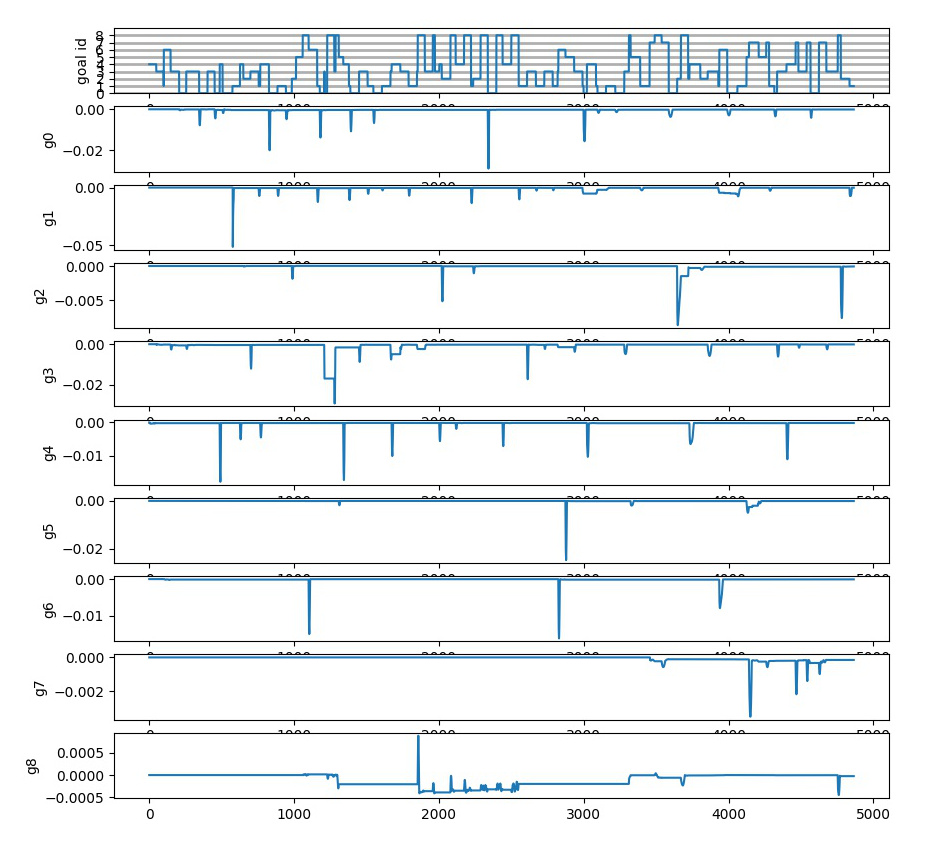}}
\caption{A sample run (experiment \#4) showing the prediction error dynamics at the goal level. The top plot shows which of the nine goals has been selected over time. The other plots show the values of the slope of the linear regression calculated on each goal buffer.}
\label{fig:goal_buffer}
\end{figure}

Figure \ref{fig:slopes} shows a more detailed view of the dynamics of the system in two sample runs of -- the worst performing -- experiment \#3 (plots on the left) and -- the best performing -- experiment \#4 (plots on the right). The diagrams show, from top to bottom: 1) the amplitude of the executed movements of the robot over time; 2) the standard deviation of the normal distribution where the exploration noise is sampled from; 3) the slope of the linear regression calculated on the MSE buffer of the forward model; 4) the maximum size of the goal error buffers. In experiment \#3, the standard deviation of the exploration noise is fixed, as it can be seen from the constant value in the second plot and the seemingly constant amplitude of the movements. The size of the goal error buffer starts increasing when the slope of the MSE dynamics becomes positive. Different trends can be observed in experiment \#4, where the exploration noise (second diagram) is adaptive and proportional to the slope of the MSE dynamics (third plot). The amplitude of the movements of the robot is also proportional to the added noise. Interestingly, around the beginning of the second half of the experiment, the MSE dynamics show an ascending trend (positive value of the slope of the MSE dynamics), likely due to an exploration of a local minimum. This makes the goal error buffer size increase, as well as the exploration noise. Eventually, the system manages to get out of the minimum, as suggested by the MSE trend, which starts decreasing again (negative value of the slope of the MSE dynamics). The general improvement of the performance makes the system liberate processing resources, reducing the size of the goal error buffers. 
Finally, Figure \ref{fig:goal_buffer} shows the slopes of the goal error buffers for the same run of experiment \#4. The plot on the top shows which goal is being selected over time, while the other plots show the slope of the error buffers for each goal. Most of the goals show initial descending trends when selected. The system maintains the goal until the slope is positive or its absolute value is below a certain threshold, or a minimum number of iterations is reached.

\section{Conclusions}
\label{sec:discussions}
We have presented a learning architecture that generates exploratory behaviours towards self-generated goals, and that regulates goal selection and the balance between exploitation and exploration through a multi-level monitoring of prediction error dynamics. The framework monitors two types of error: (1) a high-level, \textit{general error} of the system, which consists of the mean squared error (MSE) of the \textit{forward model} calculated on a pre-recorded test dataset; (2) low-level goal errors, which consist of the prediction errors that are estimated by the system when trying to reach each specific goal. 
Our architecture modulates exploration noise and leverages computational resources according to the dynamics of the overall performance of the learning system. Our results suggest that such an approach, together with dynamic goals that are learned over time, outperforms standard intrinsic motivation approaches, where exploratory noise and goals are fixed and greedy strategy is applied. More experimentation, especially in the context of more complex robotic actuators, is however required to provide further support to this claim.

In \cite{oudeyer2007intrinsic}, the authors present an algorithm showing Intelligent Adaptive Curiosity (ICA). This algorithm is capable of learning its sensorimotor space and performing goals based on the minimization of the prediction error. The algorithm divides the space in regions and attaches an expert for each region. In \cite{oudeyer2009intrinsic} the authors present a topology of computational approaches to model intrinsic motivation. They present very interesting distinctions between the different existing models. We believe that the approach presented here, as a proof of concept, can be used to function as several of the examples mentioned by the authors. To mention a couple, under the competence-based models: the monitoring of the prediction error dynamics can serve for maximizing competence by staying in goals where the error is low. At the same time, it can serve for maximizing incompetence by practicing goals where performance is at the lowest. Within the predictive models, likewise the system can choose whether to look to stay in goals where prediction error is being minimized, or search for instances where the error is high but can be lowered.

The intrinsic motivation architecture grounded on multi-level prediction error dynamics provides new insights towards the understanding of the underlying mechanisms of motivation and the emergence of emotions that drive behaviors and goal selection to promote learning. Tracking our emotions may be a necessary condition for development and cognition. The architecture presented here sets a baseline for further experimentation to shed light on the relevance of prediction error dynamics for the computational modeling and understanding of intrinsic motivation and emotion.

\bibliographystyle{./bibliography/IEEEtran}
\bibliography{./bibliography/IEEEabrv,./bibliography/biblio}

\begin{thebibliography}{10}
\providecommand{\url}[1]{#1}
\csname url@samestyle\endcsname
\providecommand{\newblock}{\relax}
\providecommand{\bibinfo}[2]{#2}
\providecommand{\BIBentrySTDinterwordspacing}{\spaceskip=0pt\relax}
\providecommand{\BIBentryALTinterwordstretchfactor}{4}
\providecommand{\BIBentryALTinterwordspacing}{\spaceskip=\fontdimen2\font plus
\BIBentryALTinterwordstretchfactor\fontdimen3\font minus
  \fontdimen4\font\relax}
\providecommand{\BIBforeignlanguage}[2]{{%
\expandafter\ifx\csname l@#1\endcsname\relax
\typeout{** WARNING: IEEEtran.bst: No hyphenation pattern has been}%
\typeout{** loaded for the language `#1'. Using the pattern for}%
\typeout{** the default language instead.}%
\else
\language=\csname l@#1\endcsname
\fi
#2}}
\providecommand{\BIBdecl}{\relax}
\BIBdecl

\bibitem{rolf2014autonomous}
M.~Rolf and M.~Asada, ``Autonomous development of goals: From generic rewards
  to goal and self detection,'' in \emph{4th International Conference on
  Development and Learning and on Epigenetic Robotics}.\hskip 1em plus 0.5em
  minus 0.4em\relax IEEE, 2014, pp. 187--194.

\bibitem{jamone2011learning}
L.~Jamone, L.~Natale, K.~Hashimoto, G.~Sandini, and A.~Takanishi, ``Learning
  task space control through goal directed exploration,'' in \emph{2011 IEEE
  International Conference on Robotics and Biomimetics}.\hskip 1em plus 0.5em
  minus 0.4em\relax IEEE, 2011, pp. 702--708.

\bibitem{rayyes2019online}
R.~Rayyes and J.~Steil, ``Online associative multi-stage goal babbling toward
  versatile learning of sensorimotor skills,'' in \emph{2019 Joint IEEE 9th
  International Conference on Development and Learning and Epigenetic Robotics
  (ICDL-EpiRob)}.\hskip 1em plus 0.5em minus 0.4em\relax IEEE, 2019, pp.
  327--334.

\bibitem{schillaci2016exploration}
G.~Schillaci, V.~V. Hafner, and B.~Lara, ``Exploration behaviors, body
  representations, and simulation processes for the development of cognition in
  artificial agents,'' \emph{Frontiers in Robotics\&AI}, vol.~3, p.~39, 2016.

\bibitem{van2017affective}
S.~Van~de Cruys, \emph{Affective value in the predictive mind}.\hskip 1em plus
  0.5em minus 0.4em\relax Johannes Gutenberg-Universit{\"a}t Mainz, 2017.

\bibitem{carver1990origins}
C.~S. Carver and M.~F. Scheier, ``Origins and functions of positive and
  negative affect: a control-process view.'' \emph{Psychological review},
  vol.~97, no.~1, p.~19, 1990.

\bibitem{lord1987control}
R.~G. Lord and P.~J. Hanges, ``A control system model of organizational
  motivation: Theoretical development and applied implications,''
  \emph{Behavioral Science}, vol.~32, no.~3, pp. 161--178, 1987.

\bibitem{o2020unraveling}
R.~C. O'Reilly, ``Unraveling the mysteries of motivation,'' \emph{Trends in
  Cognitive Sciences}, 2020.

\bibitem{von2003development}
C.~von Hofsten, \emph{On the development of perception and action}.\hskip 1em
  plus 0.5em minus 0.4em\relax London: Sage, 2003.

\bibitem{oudeyer2009intrinsic}
P.-Y. Oudeyer and F.~Kaplan, ``What is intrinsic motivation? a typology of
  computational approaches,'' \emph{Front. in Neurorobotics}, vol.~1, p.~6,
  2009.

\bibitem{parisi2019rethinking}
G.~I. Parisi and C.~Kanan, ``Rethinking continual learning for autonomous
  agents and robots,'' \emph{arXiv preprint arXiv:1907.01929}, 2019.

\bibitem{oudeyer2007intrinsic}
P.-Y. Oudeyer, F.~Kaplan, and V.~V. Hafner, ``Intrinsic motivation systems for
  autonomous mental development,'' \emph{IEEE transactions on evolutionary
  computation}, vol.~11, no.~2, pp. 265--286, 2007.

\bibitem{schillaci2020intrinsic}
G.~Schillaci, A.~P. Villalpando, V.~V. Hafner, P.~Hanappe, D.~Colliaux, and
  T.~Wintz, ``Intrinsic motivation and episodic memories for robot exploration
  of high-dimensional sensory spaces,'' in \emph{SAGE Journal on Adaptive
  Behavior (to appear)}, 2020.

\bibitem{kiverstein2019feeling}
J.~Kiverstein, M.~Miller, and E.~Rietveld, ``The feeling of grip: novelty,
  error dynamics, and the predictive brain,'' \emph{Synthese}, vol. 196, no.~7,
  pp. 2847--2869, 2019.

\bibitem{adolphs2017should}
R.~Adolphs, ``How should neuroscience study emotions? by distinguishing emotion
  states, concepts, and experiences,'' \emph{Social Cognitive and Affective
  Neuroscience}, vol.~12, no.~1, pp. 24--31, 2017.

\bibitem{rushworth2004action}
M.~Rushworth, M.~E. Walton, S.~W. Kennerley, and D.~Bannerman, ``Action sets
  and decisions in the medial frontal cortex,'' \emph{Trends in cognitive
  sciences}, vol.~8, no.~9, pp. 410--417, 2004.

\bibitem{joffily2013emotional}
M.~Joffily and G.~Coricelli, ``Emotional valence and the free-energy
  principle,'' \emph{PLoS computational biology}, vol.~9, no.~6, p. e1003094,
  2013.

\bibitem{van2011putting}
S.~Van~de Cruys and J.~Wagemans, ``Putting reward in art: a tentative
  prediction error account of visual art,'' \emph{i-Perception}, vol.~2, no.~9,
  pp. 1035--1062, 2011.

\bibitem{hsee1991velocity}
C.~K. Hsee and R.~P. Abelson, ``Velocity relation: Satisfaction as a function
  of the first derivative of outcome over time.'' \emph{Journal of Personality
  and Social Psychology}, vol.~60, no.~3, p. 341, 1991.

\bibitem{kluger1996effects}
A.~N. Kluger and A.~DeNisi, ``The effects of feedback interventions on
  performance: A historical review, a meta-analysis, and a preliminary feedback
  intervention theory.'' \emph{Psychological bulletin}, vol. 119, no.~2, p.
  254, 1996.

\bibitem{kohonen1998self}
T.~Kohonen, ``The self-organizing map,'' \emph{Neurocomputing}, vol.~21, no.
  1-3, pp. 1--6, 1998.

\bibitem{masci2011stacked}
J.~Masci, U.~Meier, D.~Cirecsan, and J.~Schmidhuber, ``Stacked convolutional
  auto-encoders for hierarchical feature extraction,'' in \emph{International
  Conference on Artificial Neural Networks}.\hskip 1em plus 0.5em minus
  0.4em\relax Springer, 2011, pp. 52--59.

\bibitem{chen2005similarity}
C.-C. Chen and H.-T. Chu, ``Similarity measurement between images,'' in
  \emph{Computer Software and Applications Conference, 2005. COMPSAC 2005. 29th
  Annual International}, vol.~2.\hskip 1em plus 0.5em minus 0.4em\relax IEEE,
  2005, pp. 41--42.

\bibitem{guido_schillaci_2020_3552827}
\BIBentryALTinterwordspacing
G.~Schillaci and A.~P. Villalpando, ``{Visuo-motor dataset recorded from a
  micro-farming robot},'' Feb. 2020. [Online]. Available:
  \url{https://doi.org/10.5281/zenodo.3552827}
\BIBentrySTDinterwordspacing

\end{thebibliography}


\end{document}